\documentclass[letterpaper, 10 pt, conference]{ieeeconf}  

\IEEEoverridecommandlockouts                              

\usepackage{amsmath} 
\usepackage{amssymb}  
\usepackage{xcolor}
\usepackage{graphicx}
\usepackage{caption}
\newcommand{\x}{\mathbf{x}}
\newcommand{\z}{\mathbf{z}}

\usepackage[ruled,vlined]{algorithm2e}
\usepackage{tabu}
\usepackage{multirow}
\usepackage{subcaption}
\usepackage{pdflscape}
\usepackage{booktabs}
\usepackage{ulem}
\usepackage{tikz}
\usepackage{todonotes}
\usepackage{soul}
\usepackage{setspace}
\usepackage{url}

\title{\LARGE \bf
VP-GO: a ``light'' action-conditioned visual prediction model 
}

\author{
        Anji Ma$^{1,2}$, Yoann Fleytoux$^2$, Jean-Bapstiste Mouret$^2$ and Serena Ivaldi$^{2}$
\thanks{*This work was partially supported by the CHIST-ERA project HEAP.}
\thanks{$^{1}$School of Mechatronical Engineering, Beijing Institute of Technology, Beijing 100081, China}
\thanks{$^{2}$ Inria, University of Lorraine, CNRS, Loria, F-54000, France }%
}

\begin{document}

\maketitle
\thispagestyle{empty}
\pagestyle{empty}

\begin{abstract}

        Visual prediction models are a promising solution for visual-based robotic grasping of cluttered, unknown soft objects. Previous models from the literature are computationally greedy, which limits reproducibility; although some consider stochasticity in the prediction model, it is often too weak to catch the reality of robotics experiments involving grasping such objects.
        Furthermore, previous work focused on elementary movements that are not efficient to reason in terms of more complex semantic actions.
        To address these limitations, we propose VP-GO, a ``light'' stochastic action-conditioned visual prediction model. We propose a hierarchical decomposition of semantic grasping and manipulation actions into elementary end-effector movements, to ensure compatibility with existing models and datasets for visual prediction of robotic actions such as RoboNet. 
        We also record and release a new open dataset for visual prediction of object grasping, called PandaGrasp.
        Our model can be pre-trained on RoboNet and fine-tuned on PandaGrasp, and performs similarly to more complex models in terms of signal prediction metrics. Qualitatively, it outperforms when predicting the outcome of complex grasps performed by our robot. 

\end{abstract}

\section{INTRODUCTION}

There are several industrial scenarios where robots have to grasp or manipulate a variety of objects from uncluttered heaps, without relying on objects models (unknown or unavailable), nor tactile/force sensing \cite{romano2011human}.
A typical scenario is in  waste sorting \cite{9395690}, where human experts plan appropriate sequences of actions to interact with complex objects.
It would be desirable to automatize the process, i.e., to 
automatically find the sequence of actions that enable the robot to pick  objects, even to grasp all the objects in the heap.

In the absence of robot sensing other than a camera and without prior knowledge on the objects, a promising method to address this problem is visual Model Predictive Control (MPC) or visual foresight \cite{ebert2018visual}. Such a technique requires a visual prediction model, i.e., a model that predicts the visual outcome (i.e., the future camera images) of the robot's action.
Because of the high dimensionality of images, visual MPC was considered unfeasible until few years ago, when video prediction models based on deep neural networks started to show very promising results in computer vision  \cite{5206848} and also in robotics \cite{DBLP:journals/corr/abs-1910-11215}. 
The first video prediction models were deterministic \cite{finn2016unsupervised}; to deal with the uncertainty caused by the interaction of the robot with the real world, several studies proposed stochastic models \cite{denton2018stochastic}, where learned prior networks catch the ``stochasticity'' .
Computationally, these methods are very greedy in terms of computation and training data.

For robotics applications, the seminal demonstration of visual MPC was done in \cite{DBLP:journals/corr/FinnL16}, showing that visual predictions could be used to plan the robot's end-effector towards desired goals. In that work and the follow-up \cite{ebert2018visual} the robot's actions were basically differential displacements of the end-effector, which was coherent with the babbling-like exploration of the robot's workspace (similar to what has been done in the last decades in developmental robotics \cite{ridge2010self}). However, for industrial applications involving repetitive grasping this action representation is not appropriate: on the one side it increases the complexity in sample-based planning to execute a complex goal-driven sequence of actions, on the other side it does not carry the semantic description of high-level actions such as grasping or picking.

To address these limitations, we propose Visual Prediction Model for Grasping Objects (VP-GO), a ``light'' stochastic action-conditioned visual prediction model. 
Our model is based on SVG-LP \cite{denton2018stochastic}, a groundbreaking work that proposed to use a learned prior model to deal with the time-varying stochasticity. In VP-GO we leverage the learned prior to catch the stochasticity of the real world grasping actions.
SVG'~\cite{DBLP:journals/corr/abs-1911-01655} and GHVAE~\cite{Wu_2021_CVPR}, both follow-ups of SVG-LP,
scaled up SVG-LP in complexity to deal with large datasets, but are very greedy in computational resources.
Here, we revisit the principle of SVG-LP. Compared to SVG-LP, our model introduces action conditioning, uses convolutional Long-Short-Term Memory networks (LSTM) instead of a simple LSTM and has a deeper architecture. Compared to SVG', which also uses convolutional LSTM, our networks have a ``deeper'' encoder, while significantly reducing at the same time the number of parameters
(5 times less than the state-of-the-art models, hence the ``light'' adjective).
Our model can compete with the existing models while being computationally efficient to train on a relatively small cluster.

Our target is grasping objects efficiently, ultimately by planning sequences of high-level grasps. To this end, we propose a hierarchical decomposition of semantic grasping and manipulation actions into elementary low-level actions (i.e., end-effector displacements).
This is frequently done in developmental learning to scaffold complex actions into elementary actions, especially for learning manipulation \cite{ivaldi2012perception}.
This kind of representation has the key advantage of producing predictive models that can be used for downstream planning of sequences of high-level grasping and manipulation actions.
Our model does not acquire semantic information explicitly (i.e., it is not an input of the network), but implicitly during the training process.
In order to train the model with rich semantic information, we acquire a new dataset (PandaGrasp) with sequences of semantic actions defined per our hierarchy, executed by a Franka robot. 
This design enables us to compare with state-of-the-art models and reuse large-scale datasets for visual prediction of robotic actions such as RoboNet \cite{DBLP:journals/corr/abs-1910-11215} to pre-train our model. 

In addition to providing the source code of VP-GO, we release PandaGrasp, the dataset recorded with our Franka setup. Differently from RoboNet, it is focused on meaningful grasping actions, and therefore can greatly help the community to study visual prediction models specific for grasping.

\section{RELATED WORK}
\textbf{Robot grasping and manipulation}
have been widely studied for many years. 
Traditionally, grasping has been studied under the umbrella of dynamic modeling and control, considering contact force and wrenches \cite{siciliano2016springer}, which require the object model or force/tactile sensing. Recent visual-based approaches are more focused on data-driven methods, using either prior knowledge like 3D models \cite{goldfeder2009columbia} or points cloud \cite{zeng2017multi} to learn objects model. These approaches are very interesting, but require the knowledge of the objects models.
Manipulation actions such as push-pulls are medium/low-level actions used to bring the environment to a more convenient configuration to solve the robot's task \cite{omrvcen2009autonomous, clavera2017policy}.
Planning the sequence of appropriate actions to solve a task is a difficult decision problem per se, and combining grasping and pushing-pulling to pick cluttered objects is a challenge \cite{dogar2012planning, boularias2015learning, zeng2018learning} . 
Several works learn the policy using model-free optimization \cite{redmon2015real, pinto2016supersizing, gualtieri2016high, mahler2017dex}, with the issue of exploring a large space.
To grasp an object that is partially occluded, or bring the objects in the workspace into a specific configuration that is more favorable to grasp an object of interest, a promising approach is to leverage visual prediction models \cite{ebert2018visual} to inform a visual MPC method. The principle consists in predicting the visual output of the robot camera after executing an action or a sequence of actions, and to use this prediction to inform a an optimal controller or decision planner.

\textbf{Visual Model Predictive Control (MPC)}
is an appealing method to address visual-based manipulation.
Traditionally, model predictive control (MPC) relies on known or learned dynamics robot models, and it is used to generate advanced motor controls. 
Once the model of the effect of the robot's actions on its environment (e.g., workspace with objects) is known or learned, it can be used to plan a sequence of optimal actions to fulfill a task: this can be done with MPC, often with a receding horizon approach, but also with reinforcement learning \cite{polydoros2017survey, massoud2009model, zhang2019solar}. For planning sequence of manipulation actions, this was often done with reduced visual models computing features of the objects in the scene, to lower the dimensionality of the problem.
With the increased availability and capabilities of computing clusters, data-driven methods and deep learning technologies \cite{goodfellow2016deep} can now be used to build models directly from high dimensional input such as images.
Visual models that leverage deep learning have been used to address visually dominant tasks such as manipulation and navigation, to learn models used for planning and decision: a visual MPC is therefore possible.
The main difficulty of dealing with visual models lies in the bigger state space, with discontinuities that make planning actions more challenging. This is particularly challenging for grasping: for example, objects disappearing from the workspace cause visual discontinuities.
In \cite{ebert2018visual}, sample-based planning based on CEM \cite{botev2013cross} was used for visual model predictive control.

\textbf{Video prediction model}
went through a significant development in the last years. Initial video/visual prediction model were based on deterministic models \cite{walker2015dense,finn2016unsupervised,xue2016visual,liang2017dual,van2017transformation,liu2017video,chen2017video,lu2017flexible}.
Their main limit is dealing with uncertainty, more precisely with the uncertain or stochastic outcomes.
Probabilistic models \cite{rasch1993probabilistic} were further proposed to carry the ``stochasticity'' information, in particular VAE-based \cite{kingma2013autoencoding} stochastic video prediction models \cite{babaeizadeh2017stochastic, denton2018stochastic, lee2018stochastic, Wu_2021_CVPR}. \cite{denton2018stochastic} proposed to use a learned prior to model time-variational stochasticity;
\cite{Wu_2021_CVPR} used a hierarchical latent variational to model the stochasticity. \cite{DBLP:journals/corr/abs-1911-01655} investigated improving performance by increasing model size.
Some other works use discrete represent of pixels \cite{Salimans2017PixeCNN, oord2016pixel} to optimize the likelihood directly through autoregressive models \cite{akaike1969fitting} and transformers \cite{vaswani2017attention, weissenborn2020scaling}. 
Autoregressive models have the advantage of producing higher quality images; the models using transformers structures can be trained in parallel and faster, although the pixel-level prediction that needs pixel by pixel reconstruction usually takes a long time and limits its use in online control tasks.

\section{Materials \& Methods}
\textbf{Robot setup:}
We use a Franka Panda robot equipped with a gripper to manipulate rigid and deformable/soft objects organized in unstructured heaps inside a plastic gray box. A Intel RealSense camera  in front of the robot is used to capture the workspace and the robot's terminal part.

\begin{figure}[t]
        \centering
        \includegraphics[width=\linewidth]{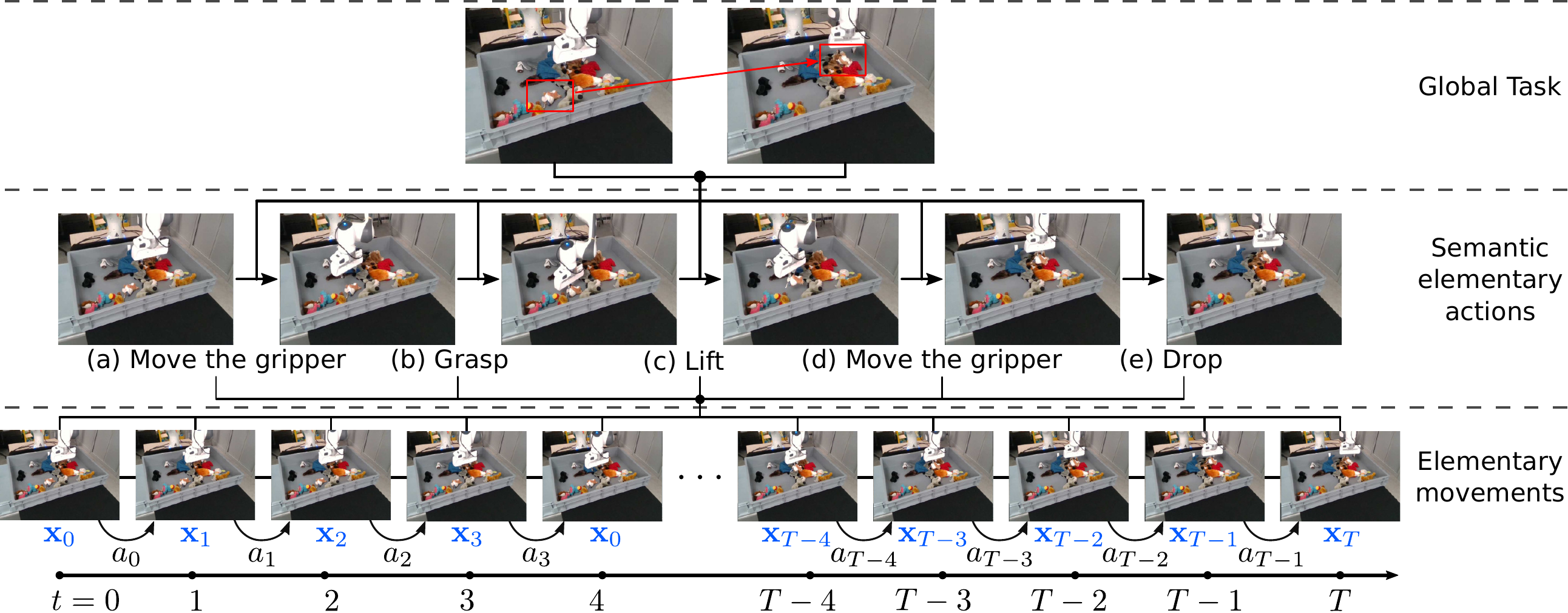}
        \caption{The hierarchical structure of decomposition semantic actions. A semantic action like sorting objects by grasping contains five element actions: (middle) (a) moving the gripper to a top position of the grasp point, (b) falling the gripper grasps the object, (c) lifting the gripper back to the top position, (d) moving the gripper to the top of the target point, (e) opening the gripper to drop the object. (bottom) Each action element is decomposed into several action movements to a sequence of images and actions.
        }
        \label{fig:grasp_action}
\end{figure}

\textbf{Datasets:}
In this paper, we use two main datasets.
\textbf{1) RoboNet}: this is a large scale dataset with more than 150K trajectories executed by 7 different robots, for a total of 15 million frames \cite{DBLP:journals/corr/abs-1910-11215}.
\textbf{2) PandaGrasp}: this is our ``smaller'' dataset recorded with our setup, with the Panda robot executing 5K trajectories for a total of 150K image frames. This is a new dataset that we are releasing as open: \url{https://gitlab.inria.fr/sivaldi/PandaGrasp_dataset}.
As in RoboNet, PandaGrasp was acquired via autonomous exploration. However, our exploration strategy was more ``efficient'' for acquiring more grasping samples.
Indeed, exploration through random actions or babbling in a large space lead to datasets with a majority of non-successful and non-purposive actions that have little to no effect on the visual appearance of the workspace. In contrast, we are interested into visual sequences where the environment significantly changes as a result of grasping actions and manipulations. Finn et al. \cite{finn2017deep} used a ``reflex'' primitive that automatically closes the gripper when it is lower than a threshold to make meaningful actions more frequent. Since we are interested into grasping, we defined an heuristic such that the robot only attempts to grasp random points sampled from the surface of the heap, extracted from the 3D point cloud: in this way, there is a small possibility of grasping in an empty area.

\textbf{
Our method: 
Action conditional stochastic Visual Prediction for Grasping Objects (VP-GO)\\}
We propose a VP-GO: a VAE-based, action conditional stochastic visual prediction model that is specific to deal with high stochasticity such as in robotic grasping  unknown objects in a heap. Our model is inspired by SVG-LP \cite{denton2018stochastic} and its successors SVG' and GHVAE, but it is ``lighter'' than those, in that it only contains simple convolutional and recurrent networks and has a smaller number of parameters to train.
Furthermore, it is an action conditional model that can deal with semantic action description. We propose a hierarchical structure to decompose high-level semantic actions into more elementary robot movements, compatible with existing models and datasets. 
Our model can be pre-trained with large-scale datasets such as RoboNet and then fine-tuned on specific smaller-scale datasets such as our PandaGrasp dataset. The code is available here: \url{https://gitlab.inria.fr/sivaldi/HEAP-VP-GO}.

In the following, we detail the two main parts of VP-GO:

\textit{1) Action decomposition -- from semantic actions to elementary movements:}\\
The action conditional visual prediction model is a deep neural network model that inputs the current and past images and generates a sequence of predicted images of the future. In general, the data used to train the model is organized into a sequence of pairs of images $(\x_0, \x_1, \ldots, \x_{T-1}, \x_T)$ and actions $(\mathbf{a}_0, \mathbf{a}_1, \ldots, \mathbf{a}_{T-1})$. 
In \cite{ebert2018visual} the action $\mathbf{a} \in \mathbb{R}^n$ is defined as a $\Delta$-displacement of the end-effector, where $n$ is the degree of freedom of the end-effector. This representation fits for describing push-pull and simple gripper movements or babbling actions. However, it is very limited to describe more complex manipulations.
In contrast, in our application the robot should purposefully interact with objects, executing many grasps: our ultimate goal is to find sequences of grasping actions to execute on the objects in the heap.
For this reason, we use not only low-level displacement actions but also high-level semantic actions such as grasping and picking. 
In order to integrate semantic actions into a generic prediction model, while being compatible with previous video prediction models and corresponding datasets (e.g., RoboNet), we propose a hierarchical structure of high-level semantic robot actions. High-level actions such as ``grasp'' can be decomposed into several elementary movements that can directly be trained in a general model, as shown in Fig. \ref{fig:grasp_action}. We split a semantic grasping action into several action elements: moving the gripper to a top position over the grasp point, lowering the gripper to the grasp point and closing it, lifting the gripper back to the top position, moving the gripper to the top of the target point, and opening the gripper to drop the object. Finally,  we decompose each elementary action into several elementary movements that are defined as a displacement of the end effector's elements.

\begin{figure*}[t]
        \centering
        \includegraphics[width=0.7\linewidth]{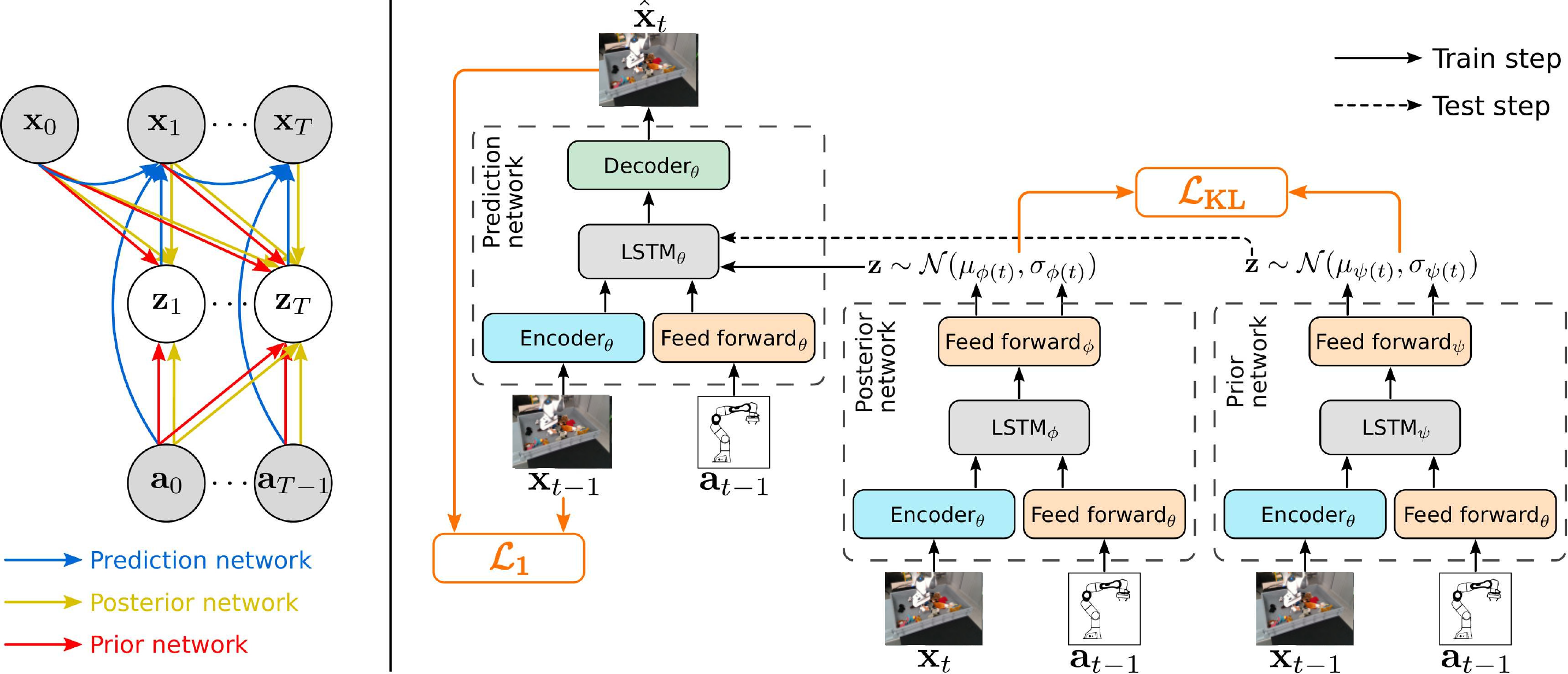}
        \caption{\textit{Left}: Probabilistic model of VP-GO.
                Blue, yellow and red lines indicate the prediction, posterior an prior models respectively.
                \textit{Right}: Detailed structure of VP-GO. Each network contains deep convolutional layers as the encoder for images, feed forward network for actions and LSTM layers to condition the previous input. Prior network and posterior network contain a feed forward network which outputs the mean and variance of the latent variables after the LSTM layers. In the prediction network, deep convolutional layers (decoders) are used to generate prediction images.
        }
        \label{fig:sto}
\end{figure*}
      
\textit{2) Action conditioned stochastic visual prediction model: }\\
As shown in the left part of Fig. \ref{fig:sto}, the stochastic prediction model takes $c$ visible frames $\x_0, \x_1, \ldots, \x_c$ and an action sequence $\mathbf{a}_0, \mathbf{a}_1, \ldots, \mathbf{a}_{T-1}$ to predict several futures $\x_{c+1}, \ldots, \x_{T-1}, \x_T$. 
To deal with the stochastic nature of the real world, the VAE-based prediction model introduces latent variables $\z \sim p(\z)$ that carries the stochastic information. We can generate the image $\x_t$ at time $t$ from a prediction model $p_\theta(\x_{t}|\x_{0:t-1}, \mathbf{a}_{0:t-1}, \z_{1:t})$ conditioned on the previous images $\x_{0:t-1}$, actions $\mathbf{a}_{0:t-1}$ and the latent variables $\z_{1:t}$.
\\
Since $p_\theta(\x_t)$ can not be directly maximized over the distribution of $p(\z_t)$, the VAE-based method approximates the posterior $p(\z_t|\x_{0:t-1}, \mathbf{a}_{0:t-1}, \z_{1:t-1})$ by an inference network that is parametrized as a conditional Gaussian  distribution $q_{\phi}(\z_{t}|\x_{0:t}, \mathbf{a}_{0:t-1}, \z_{1:t-1}) = \mathcal{N}(\mu_\phi(\x_{0:t}, \mathbf{a}_{0:t-1}, \z_{1:t-1}),\sigma_\phi(\x_{0:t}, \mathbf{a}_{0:t-1}, \z_{1:t-1}))$.

Previous work \cite{babaeizadeh2017stochastic} assumed that $p(\z_t)$ is a fixed Gaussian distribution $\mathcal{N}(0, \mathbf{I})$. In contrast, to catch the rich stochasticity that is caused in our application by the robot's grasping action, 
we assume that the learned-prior $p(\z_t)$ as in \cite{denton2018stochastic} varies across time, conditioned on actions and previous images. Specifically, the learned prior is also parameterized as a conditional Gaussian distribution as  $p_\psi(\z_{t}|\x_{0:t-1}, \mathbf{a}_{0:t-1}, \z_{1:t-1}) = \mathcal{N}(\mu_\psi(\x_{0:t-1}, \mathbf{a}_{0:t-1}, \z_{1:t-1}),\sigma_\psi(\x_{0:t-1}, \mathbf{a}_{0:t-1}, \z_{1:t-1}))$.
We train the entire model by maximizing the variational lower bound of the log-likelihood as in the variational autoencoders:
\begin{equation}                 
    \mathcal{L}_{\theta,\phi,\psi}(\x_{c+1:T}) =
    \sum_{t=c+1}^{T} E \, LP
  - \sum_{t=c+1}^{T} D_{KL} (QP|| PP)
  \label{equ:2}
\end{equation}
where $E=\mathbb{E}_{q_{\phi}(\z_{t}| \x_{0:t}, \mathbf{a}_{0:t-1}, \z_{1:t-1})}$, \\
$LP= \log p_\theta(\x_{t}|\x_{0:t-1}, \mathbf{a}_{0:t-1}, \z_{1:t})$, \\
$QP=q_{\phi}(\z_{t} | \x_{0:t}, \mathbf{a}_{0:t-1}, \z_{1:t-1})$, \\
$PP=p_\psi(\z_{t}|\x_{0:t-1}, \mathbf{a}_{0:t-1}, \z_{1:t-1})$.

We parametrize the prediction network with a fixed-variance Laplace distribution $\hat{\x}_{t} \sim \text{Laplace}(b_\theta(\x_{0:t-1}, \mathbf{a}_{0:t-1}, \z_{1:t}), \sigma)$; in other words, we reduce the first term on the right hand side to $\ell_1$ loss instead of $\ell_2$ loss between $\hat{\x}$ and $\x$ to help generate sharper images \cite{DBLP:journals/corr/abs-1911-01655}.
During training, the latent variables $\z_{t}$ are sampled from posterior $q_{\phi}(\z_t)$ that inputs the ground truth $\x_{t+1}$, and the second term on the right hand side is a KL-divergence that fits the posterior $q_{\phi}(\z_t)$ to the learned-prior $p_{\psi}(\z_t)$. During testing, since the ground truth is unavailable, we directly sample $\z_t$ from the learned prior $p_{\psi}(\z_t)$.

\textbf{Network Stucture: } 
As shown in Fig. \ref{fig:sto} (right), we use deep encoder with 16 vgg layers to decrease the spatial size and increase the channel dimensions of images from 48×64×48 pixels to 3×4×512, and a single dense feed forward network to encode actions and reshape it to 3×4×2. The encoded actions and latent variables are appended to encoded images along the channel dimension. The dependency with previous inputs is achieved by the recurrent layers. For the prediction network, we use 2 convolutional LSTM layers. For prior and posterior networks, we use one convolutional LSTM layer. The latent variables are output from a dense layer following the LSTM and the predicted image finally can be generated through a decoder that has similar structure as the encoder.

\textbf{Metrics for evaluation:}
As done in prior work~\cite{babaeizadeh2017stochastic,denton2018stochastic,DBLP:journals/corr/abs-1911-01655, castrejon2019improved,Wu_2021_CVPR}, we evaluate video prediction models across 4 metrics:
Structural Similarity Index Measure (SSIM) \cite{wang2004image}, Peak Signal-to-noise Ratio (PSNR)) \cite{huynh2008scope}, Learned Perceptual Image Patch Similarity (LPIPS) \cite{zhang2018unreasonable} and Fréchet Video Distance (FVD) \cite{unterthiner2019fvd}.
To compare PSNR, SSIM, LPIPS, \cite{babaeizadeh2017stochastic, denton2018stochastic} sample 100 rollouts for each sample and select the best trajectory.
FVD uses instead all the 100 samples \cite{DBLP:journals/corr/abs-1911-01655}.
In contrast with prior work only reporting the best score among the 100 samples, we also report the average score: the reason is that we want a better idea of the real performance of the model that will be used for downstream planning (where only one sample is used, if we refer to~\cite{ebert2018visual})
to take decisions on the next action to perform. 
We  adopted the specific settings of \cite{Wu_2021_CVPR}: we generate predictions conditioned on 2 context frames and a sequence of actions and evaluate on a 10 rollout horizon; since FVD metric can be significantly disturbed by the different batch sizes, we use batch size of 256 for FVD.

\section{EXPERIMENTS}
The experiments are designed to answer to the questions:\\
\textbf{Q1}: Can we achieve state-of-the-art performance with our ``lighter'' model VP-GO?\\
\textbf{Q2}: Can we improve the prediction performance by using a deeper network?\\
\textbf{Q3}: Can we improve the prediction performance by considering the robot's state (precisely, the end-effector position) as additional input in the network and actions?\\
\textbf{Q4}: Can the movement segments decomposed from the semantic actions be directly trained in the general video prediction model?\\
\textbf{Q5}: Can we improve the performance with fine-tuning, i.e., by leveraging a pre-trained model, instead of training a model from scratch, since we only have a small dataset?

\begin{table}[t]
  \centering
  \caption{A1: Comparison with state-of-the-art models (mean $\pm$ standard error)}
  \label{tab:test_1}
  \scriptsize
  \begin{tabular}{cccccc}
    \toprule
    \multirow{2}{*}{Dataset} & 
    \multirow{2}{*}{Model}& \multicolumn{4}{c}{Video Prediction Performance (Test)} \\
    & & FVD $\downarrow$ & PSNR $\uparrow$ & SSIM $\uparrow$ & LPIPS $\downarrow$ \\
    \midrule
    \multirow{3}{*}{\shortstack{RoboNet}} & GHVAEs
    & \textbf{95.2$\pm$2.6} & 24.7$\pm$0.2 & \textbf{89.1$\pm$0.4} & 0.036$\pm$0.001 \\
    & SVG' & 123.2$\pm$2.6 & 23.9$\pm$0.1 & 87.8$\pm$0.3 & 0.060$\pm$0.008 \\
    & Ours & 111.5±1.3 & \textbf{26.9±0.2} & \textbf{89.1±0.3} & \textbf{0.030±0.001} \\
    \bottomrule
  \end{tabular}
\end{table}

\begin{table}[t]
    \centering
    {
      \caption{A2: Ablation study of the encoder and decoder networks depth (mean $\pm$ standard error)}
      \label{tab:vgg_depth}
      \scriptsize
      \begin{tabular}{cccccc}
      \toprule
      \multirow{4}{*}{Dataset} & \multicolumn{1}{c}{\multirow{2.5}{*}{Layer config.}} & \multicolumn{4}{c}{Video Prediction Performance (Test)} \\
      \cmidrule{3-6}
      & & \multirow{1}{*}{FVD $\downarrow$} & 
      \multicolumn{1}{c}{PSNR $\uparrow$} & \multicolumn{1}{c}{SSIM $\uparrow$} & \multicolumn{1}{c}{LPIPS $\downarrow$}\\
      \midrule
      \multirow{5.1}{*}{\shortstack{RoboNet}} & \shortstack{VGG16\\layer conv3\_3} & {232.8$\pm$1.9} & 24.5±0.3 & 86.8±0.3 & 0.078±0.001\\
       & \shortstack{VGG16\\layer conv4\_3} & {129.3$\pm$1.5} & 26.6±0.2 & 88.7±0.3 & 0.035±0.001 \\
       & \shortstack{VGG19\\layer conv4\_4} & \textbf{111.5$\pm$1.3} & \textbf{26.9±0.2} & \textbf{89.1±0.3} & \textbf{0.030±0.001}\\
      \bottomrule
  \end{tabular}
    }
  \end{table}

\begin{table*}[t]
  \centering
  \caption{A3: Contribution of the robot’s state (mean $\pm$ standard error)}
  \label{tab:test_2}
  \scriptsize
  \begin{tabular}{cccccccccc}
    \toprule
    \multirow{4}{*}{Dataset} & \multicolumn{2}{c}{\multirow{2.5}{*}{Conditional Input}} & \multicolumn{7}{c}{Video Prediction Performance (Test)} \\
    \cmidrule{4-10}
     & & & \multirow{2}{*}{FVD $\downarrow$} & \multicolumn{2}{c}{PSNR $\uparrow$} & \multicolumn{2}{c}{SSIM $\uparrow$} & \multicolumn{2}{c}{LPIPS $\downarrow$} \\
    & Action & State  & & Best & Average & Best & Average & Best & Average\\
    \midrule
    \multirow{3.7}{*}{RoboNet} & $\bullet$ &  & \textbf{111.5$\pm$1.3} & 26.9±0.2 & 25.0±0.2 & 89.1±0.3 & 85.9±0.4 & 0.030±0.001 & 0.039±0.001 \\
    & $\bullet$ & $\bullet$ & 116.8$\pm$1.0 & \textbf{27.4$\pm$0.2} & \textbf{26.1$\pm$0.2} & \textbf{89.9$\pm$0.3} & \textbf{87.9$\pm$0.4} & \textbf{0.029$\pm$0.001} & \textbf{0.036$\pm$0.001} \\
    \cmidrule{2-10}
    & \multicolumn{2}{c}{Improvement $\uparrow$} & - - & $+$0.5 & \textbf{$+$1.1} & $+$0.8 & \textbf{$+$2.0} & $+$0.001 & \textbf{$+$0.003} \\
    \bottomrule
   \end{tabular}
\end{table*}

\textbf{A1: Comparing our ``lighter'' model VP-GO with GHVAE and SVG'}
We compare our ``lighter'' model VP-GO with the ``heavier'' GHVAE \cite{Wu_2021_CVPR} and SVG' \cite{DBLP:journals/corr/abs-1911-01655}. GHVAE is a hierarchical model; 
SVG' is similar to our model but with shallow convolutional layers for encoder and decoder. Both have a larger number of parameters than ours: 599 million for GHVAE, 298 million for SVG’, and 129 million for ours. Since their code is not available, to compare the performance on RoboNet we directly use the results reported in \cite{Wu_2021_CVPR}.
Table \ref{tab:test_1} shows the performance of the video prediction model on the RoboNet dataset. Our model outperforms SVG' vastly and achieves a comparable performance to GHVAE; specifically, we exceed GHVAE on both structured metric PSNR and human perceptual metric LPIPS and get similar performance on SSIM but behind on FVD.
The video attachment shows more examples of RoboNet, with several rollouts predicted by ours and the other models \cite{Wu_2021_CVPR}.
Our results suggest that a lighter model can have substantial improvement of performance by adjusting the internal structure, e.g., using deeper convolutional layers. Also, the drastic reduction of the number of parameters (from 599 to 129 millions) did not lower the performance and at the same time enabled us to run the training on a smaller cluster.
\footnote{Training our model for 400k steps on a 11GB GPU cluster (4 $\times$ GeForce RTX 2080 Ti) takes 4.7 days. We could not re-implement and re-train GHVAE as reported in \cite{Wu_2021_CVPR} since the paper reports using 24GB/48GB GPU machines.}

\textbf{A2: Ablation study of the encoder and decoder networks depth}
Compared with SVG' \cite{DBLP:journals/corr/abs-1911-01655}, one notable difference is that we use a deeper network configuration for encoder and decoder networks. 
In this section experiment, we report an ablation study to show whether using a deeper VGG \cite{simonyan2014very} layer configuration helps. As shown in Table \ref{tab:vgg_depth}, we train our models in three different configurations: VGG16 up to layer conv3\_3, VGG16 up to layer conv4\_3, and VGG19 up to layer conv4\_4 which is the default of our model. 
The deeper model VGG19 outperforms the others.
The results confirm the observation of \cite{simonyan2014very} that ``deeper models might be beneficial for larger datasets'' and support the hypothesis in \cite{Wu_2021_CVPR} that deep models with compressed height and width reduce the spatial correlations.

\begin{table*}[t]
  \centering
    \caption{A4, 5: The comparison between models trained from scratch and fine-tuning by FVD and averaged scores of PSNR, SSIM, and LPIPS on the test set of semantic action. (mean $\pm$ standard error)}
    \scriptsize
    \begin{tabular}{ccccccccccc}
    \toprule
    \multirow{3.5}{*}{Dataset} &
    \multirow{3.5}{*}{Stage}
    & \multicolumn{8}{c}{Video Prediction Performance (Test)} \\
    \cmidrule{3-10}
    & & \multicolumn{2}{c}{PSNR $\uparrow$} & \multicolumn{2}{c}{SSIM $\uparrow$} & \multicolumn{2}{c}{LPIPS $\downarrow$} & \multicolumn{2}{c}{FVD $\downarrow$}\\
    & & from scratch & fine-tuning & from scratch & fine-tuning & from scratch & fine-tuning & from scratch & fine-tuning\\
    \midrule
    \multirow{5.3}{*}{\shortstack{PandaGrasp}}
     & Approaching &\textbf{34.2$\pm$0.2} & 33.9$\pm$0.2   &\textbf{98.7$\pm$0.1} & 98.6$\pm$0.1   & 0.004$\pm$0.001 & 0.004$\pm$0.001 & \multirow{5.3}{*}{\shortstack{110.9$\pm$3.9}} & 
    \multirow{5.3}{*}{\shortstack{110.9$\pm$4.0}} \\
    & Grasping     &31.0$\pm$0.2 & 31.0$\pm$0.2   &\textbf{97.1$\pm$0.1} & 97.0$\pm$0.1   & 0.009$\pm$0.001 & 0.009$\pm$0.001 &\\
    & Moving       &25.3$\pm$0.2 & \textbf{25.9$\pm$0.2}   &90.8$\pm$0.1 & \textbf{91.3$\pm$0.1}   &0.034$\pm$0.002 & \textbf{0.032$\pm$0.002} & \\
    \cmidrule{2-8}
    & Average      &30.1$\pm$0.2 & 30.1$\pm$0.2   &95.5$\pm$0.2 & \textbf{95.6$\pm$0.2}   &0.016$\pm$0.001 & \textbf{0.015$\pm$0.001} & \\
    & Final Goal   &23.7$\pm$0.4 & \textbf{23.9$\pm$0.3}   &87.0$\pm$0.7 & \textbf{87.7$\pm$0.6}   &0.050$\pm$0.004 & \textbf{0.047$\pm$0.004} & \\
    \bottomrule
\end{tabular}
  \label{tab:Finetuning}
\end{table*}

\textbf{A3: Ablation study: Contribution of the robot's state}
A large part of the stochasticity comes from estimating the position of the robot in the real world through images.
However, the robot's state, in particularly the position of the end-effector in its workspace is easily accessible. We posit that it could help that estimation. In this experiment, we do an ablation study to show whether adding the robot's state to the action conditional model helps.
We evaluate the contribution of the robot's state on both the best performing model  and on the average score from 100 samples models.
Results are shown in Table \ref{tab:test_2}.
In both cases, using the robot state as an additional input improves the model's performance. Furthermore, using the robot state as an additional input can improve much more in the average score than in the best score. PSNR improves 1.1 in average score case than 0.5 in the best score; SSIM improves 2.0 in average score than 0.8 in the best score, LPIPS improves 0.003 in one shot case than 0.001 in the best score.
These results also confirm the intuition from \cite{Wu_2021_CVPR} that a large part of the uncertainty comes from estimating the robot's position from images.

\begin{figure}[t]
    \centering
    \includegraphics[width=0.975\linewidth]{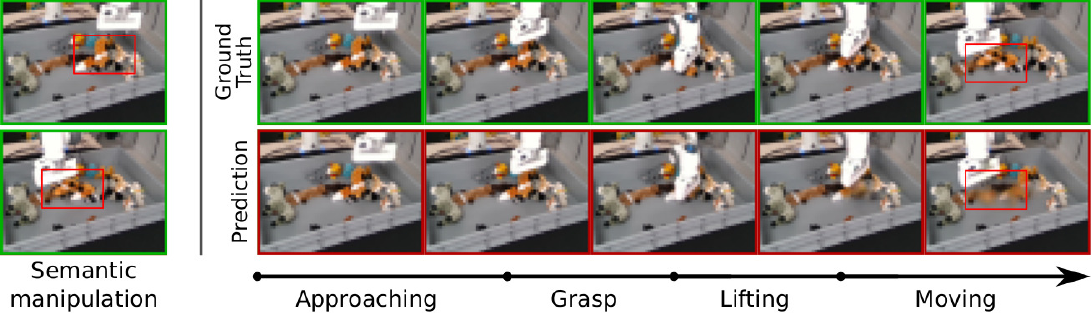}
    \caption{A4: Visual inspection of future frames predicted by VP-GO from semantic actions. The model correctly predicts the orange plush toy being grasped and moved.}
    \label{fig:result1}
  \end{figure}

\begin{figure}
    \centering
    \includegraphics[width=0.975\linewidth]{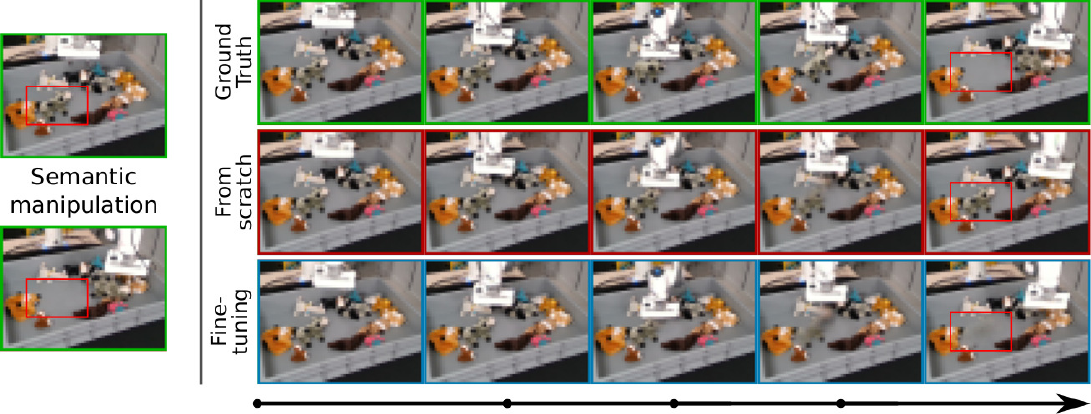}
    \caption{A5: Visual comparison of our model VP-GO trained on PandaGrasp from scratch with a VP-GO pre-trained on RoboNet then fine-tuned on PandaGrasp.}
    \label{fig:result2}
  \end{figure}

\textbf{A4: Our visual prediction model with semantic actions }
We investigate the performance of using our VP-GO for different semantic actions in Fig. \ref{fig:result1} and~\ref{fig:result2} and Table \ref{tab:Finetuning}.
Fig.~\ref{fig:result1} shows an example of prediction of an entire trajectory after the robot picks and moves an orange soft object. We can successfully predict the object's movement, compared to the ground truth.
However, when the robot is approaching the target object, the metrics in Fig.~\ref{fig:result2} and Table \ref{tab:Finetuning} show that 
there is almost no performance loss in the predicted frames; when the robot begins to grasp and interacts with the objects, 
the performance decreases, and then keep decreasing linearly with the number of frames we look ahead in the future.
This is likely due both to the higher stochasticity given by the grasping action, and the intrinsic higher uncertainty about predicting an action in the far future and may be alleviated by reducing the prediction horizon in semantic action-based planning.
With such results, VP-GO shows promising potential to be used 
for planning sequences of robot manipulation actions in the semantic action space, which is much more efficient than using sample-based planning in the movement action space. 
Still, we observe that a precise visual prediction of a moving soft object is still challenging.

\begin{figure}[t]
  \centering
  \includegraphics[width=\linewidth]{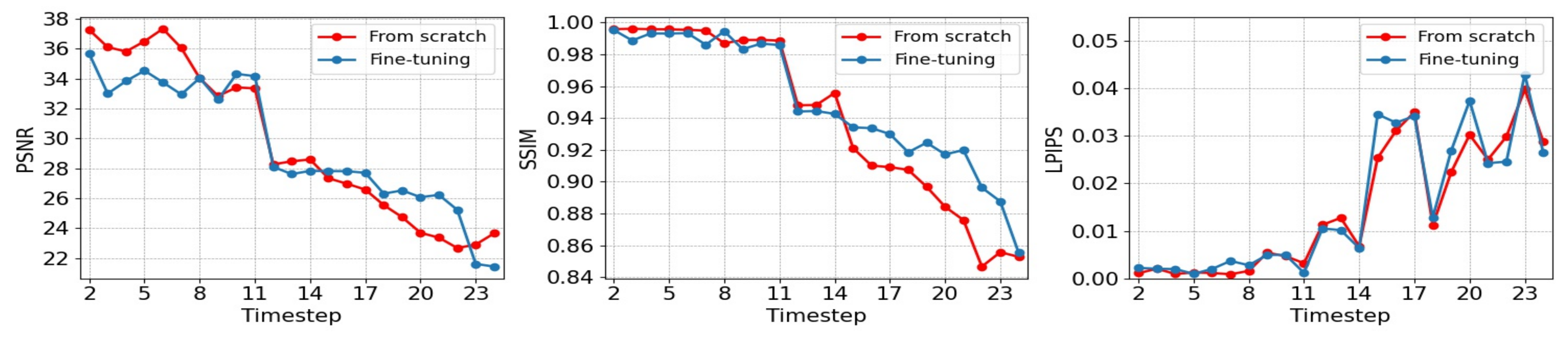}
  \caption{A5: An example of the limitation of the metrics for evaluating the performance in terms of grasping task. In terms of all three metrics, the model performance equivalent that contrary to the comparison in Fig.~\ref{fig:result2} in terms of a grasping task.}
  \label{fig:test5}
  \vspace{-0.2cm}
\end{figure}

\textbf{A5: Finetuning with PandaGrasp}
Recording large scale datasets such as RoboNet~\cite{DBLP:journals/corr/abs-1910-11215} is very expensive in terms of time and resources. Our dataset PandaGrasp is ``smaller'' in comparison, yet for many practical applications we want our models to work with datasets of this size. Hence, we want to investigate if a model trained with a ``smaller'' dataset is performing as well as a model trained with a ``bigger'' dataset and fine-tuned on the ``smaller''.
Also, PandaGrasp is more purposeful for semantic grasping actions compared to the low-level actions (for babbling-like exploration) in RoboNet. Our experiment also verifies whether our method can let prior low-level action knowledge benefits the training on more purposeful data with semantic actions.
Specifically, we compare our model VP-GO trained on PandaGrasp from scratch with a VP-GO pre-trained on RoboNet then fine-tuned on PandaGrasp. 
As shown in Fig.~\ref{fig:result2} and Table~\ref{tab:Finetuning}, 
the two models are performing in the same way, from the point of view of the metrics FVD, PSNR, SSLM and LPIPS.
However, qualitative inspection of the visual predictions performed by the two models actually shows that the fine-tuned model can correctly predict missing and misplaced objects after, while the scratch model fails.
One typical problem is failing at predicting objects that move from the background and leave empty space:
Fig.~\ref{fig:result2} shows an example where an object is grasped and moved from the initial location: the object is still in the same place in the prediction of the first model learned from scratch, while the fine-tuned model correctly predicts that the object is moved away. Fig.~\ref{fig:test5} shows the corresponding performance metrics.
Our intuition is that training on a big dataset such as RoboNet produces a more ``expert'' model, and fine-tuning on our smaller dataset PandaGrasp just makes sure that the predicted frames are more coherent with our specific setup.

\begin{figure}
\centering
  \includegraphics[width=\linewidth]{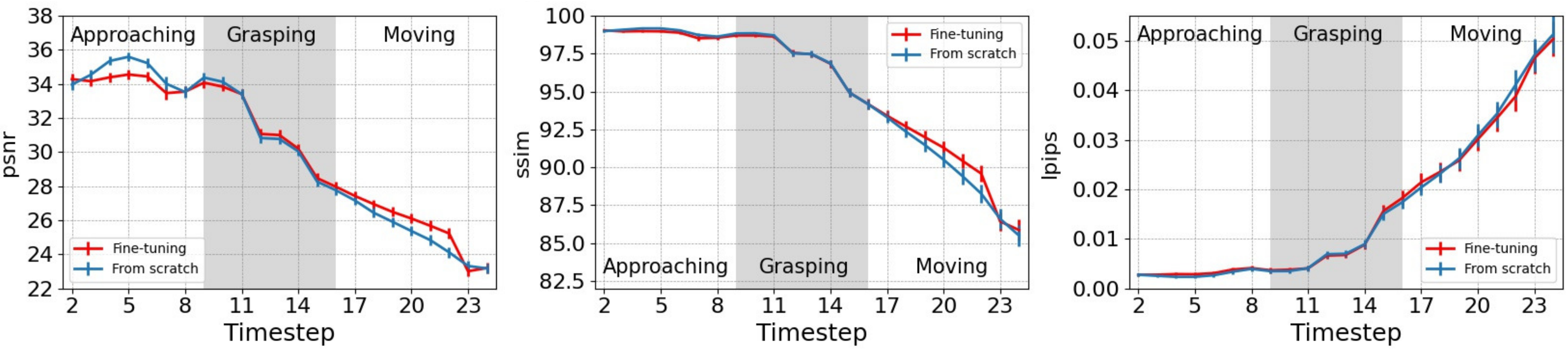}
  \caption{A4, 5: The comparison between models trained from scratch and fine-tuning by averaged scores of PSNR, SSIM, and LPIPS on the test set of semantic action.}
  \label{fig:result3}
\end{figure}

\section{ CONCLUSION}
We proposed VP-GO, a stochastic action-conditioned visual prediction model that can be used to predict future visual outcomes after grasping and manipulating irregular cluttered objects. 
Our model is inspired by SVG-LP \cite{denton2018stochastic}, 
but it introduces action conditioning, uses convolutional LSTM and has a deeper structure.
With respect to the state-of-the-art models \cite{DBLP:journals/corr/abs-1911-01655, Wu_2021_CVPR}, VP-GO is lighter in that it has considerably less parameters; still it has comparable performance.
This is promising, since lighter models are more sustainable and reproducible.
We also contributed with a new open dataset ``PandaGrasp'' that contains 5K sequences of robot grasping trajectories, executed by a Franka Panda robot, according to a proposed hierarchical decomposition of semantic actions into elementary movements.
In our experiments, 
we found that pre-training our model with a large-scale dataset (RoboNet) then fine-tuning the model with our smaller dataset PandaGrasp, specific to our robotics setup, gave better results in terms of predicting future frames after grasping and moving objects.

The next step is to use our visual prediction model in a RL framework to plan a sequence of grasps to selectively remove some objects of interests from a heap of cluttered objects.
Interestingly, our model can benefit of high-level semantic actions, reducing the complexity of downstream planning. 
VP-GO does not yet explicitly uses the semantic action label, but leverages the implicit semantic information carried out by the PandGrasp dataset during training. In the future, we plan to explicitly condition the model on the additional semantic information.








\bibliographystyle{./IEEEtran} 
\bibliography{./IEEEabrv,./root}

\end{document}